# Detail-Preserving Pooling in Deep Networks

Faraz Saeedan[1]   Nicolas Weber[1,2][*]   Michael Goesele[1,3][*]   Stefan Roth[1]

[1]TU Darmstadt   [2]NEC Laboratories Europe   [3]Oculus Research

## Abstract

*Most convolutional neural networks use some method for gradually downscaling the size of the hidden layers. This is commonly referred to as pooling, and is applied to reduce the number of parameters, improve invariance to certain distortions, and increase the receptive field size. Since pooling by nature is a lossy process, it is crucial that each such layer maintains the portion of the activations that is most important for the network's discriminability. Yet, simple maximization or averaging over blocks,* max *or* average pooling, *or plain downsampling in the form of* strided convolutions *are the standard. In this paper, we aim to leverage recent results on image downscaling for the purposes of deep learning. Inspired by the human visual system, which focuses on local spatial changes, we propose* detail-preserving pooling *(DPP), an adaptive pooling method that magnifies spatial changes and preserves important structural detail. Importantly, its parameters can be learned jointly with the rest of the network. We analyze some of its theoretical properties and show its empirical benefits on several datasets and networks, where DPP consistently outperforms previous pooling approaches.*

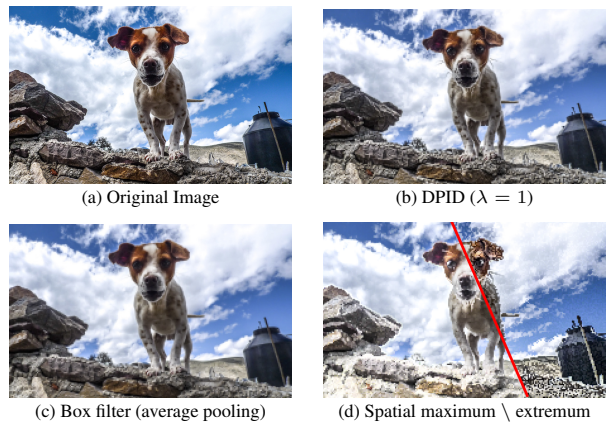

Figure 1. Visual comparison for downscaling by a factor of 16 for several methods applied to the image depicted in (a): (b) detail-preserving image downscaling (DPID) [31] with $\lambda = 1$; (c) box filter plus downsampling (average pooling); and (d) taking the maximum \ extremum in each local neighborhood (max \ extremum pooling). The details are better preserved by DPID, *e.g.*, observable in the dog's eyes and whiskers. *Best viewed on screen.*

## 1. Introduction

Pooling in various forms has been a crucial component of virtually all convolutional neural networks (CNNs) [1]. It reduces the computational cost and the number of parameters, improves the invariance to minor distortions, and increases the receptive field of intermediate and output nodes. In mid-sized networks, such as VGG [27] or GoogLeNet [29], *max* or *average pooling* are most commonly used. Deeper network architectures, such as ResNet [11], often use *strided convolutions*. All of them have shortcomings: Strided convolutions simply pick one node in a fixed position in each local neighborhood, regardless of the significance of its activation. From an image downscaling perspective, such downsampling can cause artifacts such as aliasing. Average pooling results in a gradual, constant attenuation of the contribution of individual nodes in the forward and backward passes, also irrespective of the importance of the local structure. The resulting loss of detail becomes apparent when visualizing its effect in image downscaling (Fig. 1c). Since detail is important for the discriminability of the network [1], max pooling aims to preserve it. Yet, it disrupts the flow of gradients in some of the branches of the backward pass, since only one node is selected in every input neighborhood. Moreover, looking from an image downscaling perspective, max pooling leads to implausible looking results (Fig. 1d). We thus ask: Can we design pooling layers that address these limitations by taking inspiration from image downscaling, expecting improved results?

Another shortcoming of standard pooling layers is that the question of which one performs best depends on the specific combination of network and dataset. Determining the best pooling strategy is thus often done using exhaustive empirical testing. There have been a few attempts to introduce more generic pooling layers, *e.g.*, based on learning a linear combination of max and average pooling [19], or using binary switching variables deciding between the two [32]. One drawback is that these methods inherit some of the limitations of their components, *i.e.* max and average pooling. Furthermore, a linear mixture of baselines [19] only increases the flexibility in a limited fashion, raising the question of the potential benefits of nonlinear combinations. Yet, existing nonlinear pooling layers require brute-force parameter search [2, 9] and their benefit for deep networks and/or large-scale tasks remains unclear.

---

[*]This work was carried out while at TU Darmstadt.

In this paper we aim to fill this gap. While traditional image downscaling techniques follow well-known ideas from signal processing, recent work in the graphics community has shown that downscaling yields subjectively better results if special care is taken to preserve details in the images [31] (Fig. 1b). Inspired by this observation, we first introduce a novel pooling layer that is parameterized to adjust the level of detail being preserved. Our approach is able to *learn* a suitable pooling from a *continuum of methods* that ranges from average to max (or extremum[1]) pooling. Second, we show theoretically and empirically that our method provides a *generic pooling layer* since it can learn to behave like max or average pooling, or provide a nonlinear combination of the two. Third, our approach can be combined with *stochastic regularization techniques* [34]. Fourth, our pooling layer is completely *differentiable*, which can provide benefits such as in very deep networks. Exhaustive experiments on CIFAR10 show that the proposed detail-preserving pooling (DPP) learns to perform at least as well as the best standard pooling layer for various network types, and also *outperforms* other recent pooling methods in *all settings* considered. For large-scale real-world tasks in which the pooling specifics become more important, we show that replacing the respective standard method with DPP yields improved results on the ImageNet classification task. For the popular ResNet architecture [11], we find that *ResNet-101 using DPP outperforms a plain ResNet-152* despite significantly fewer layers and parameters.

## 2. Related Work

Even before the widespread use of CNNs, pooling was used in the majority of feature extractors to reduce the size of the feature vectors and gain invariance to small transformations of the input. This is often motivated by the study of complex cells in animal visual cortex [13]. Popular uses include SIFT [21] and HOG [4], which aggregate the orientation of gradients in a neighborhood. In convolutional neural networks, max pooling [17, 18] and average pooling [15, 22] are most commonly used. In some networks, especially very deep ones such as ResNet [11], strided convolutions are used for pooling, which are efficient but not adaptive to the data. From an image downscaling perspective, they just perform a sparse but regular downsampling.

Boureau *et al*. [1] analyzed pooling methods and showed that max pooling improves discriminability over average pooling, particularly for features with low activation probability. Hence, the optimal pooling for a feature map might be somewhere 'between' average and max pooling. The proposed DPP not only yields a suitable parameterization that nonlinearly bridges max and average pooling, but also *learns the shape of this nonlinear function* for *every feature map* based on the training data and the activations thereon, enabling DPP to explore a continuum of poolings.

Earlier work has focused on altering the receptive field of a pooled pixel or on unconventional pooling ratios. Ionescu *et al*. [14] provide a methodology that enables incorporating higher-order pooling layers in deep networks. Maxout [7] suggests performing inter-channel max pooling. Fractional pooling [8] uses a fractional downscaling ratio, and hence a more gradual size reduction. Rippel *et al*. [24] downsample feature maps in spectral space using low-pass filtering. This smoothes the input rather than preserving details, which are mostly concentrated in higher frequencies.

Recent work has aimed to preserve the most discriminative aspects of the data / activations and discard the redundant ones through learnable pooling layers. Various attempts to bridge max and average pooling have been made. Mixed pooling [32] learns to perform a hard switch between max or average pooling in every layer. $L_p$ pooling [2, 9] outputs the spatial $L_p$ norm of the neighborhood, with the $L_1$ norm performing averaging (on non-negative inputs) and the $L_\infty$ norm acting like max pooling. Lee *et al*. [19] combine max and average pooling using a learned tree, choosing the desired pooling based on the input data. Motivated by regularization rather than detail preservation, S3pool [34] and stochastic pooling [33] stochastically select a node in a neighborhood, with the latter favoring stronger activations. Their benefit stems from joint pooling and regularization and it is not trivial to assess each aspect independently.

Almost all of these pooling methods rely on directly combining previous ones such as max or average pooling. This may mitigate harmful effects of the weaker baseline, but often only to a certain extent, and frees the practitioner from tediously choosing the best pooling layer for every network/dataset combination.

Other pooling concepts include scaling proportionally to the input size [5, 10, 23] such that networks can cope with varying image sizes. Global average pooling [20] is a different application of downscaling, which spatially averages the feature maps of the last convolutional layer before feeding them into the classifier.

Here, we focus on downscaling inside the network and propose a pooling layer that is trainable, fully differentiable, and includes major pooling techniques as special (limit) cases. Moreover, we build on a detail-preserving image downscaling approach [31] that outperforms conventional image downscaling techniques in subjective testing.

---

[1]By extremum pooling we refer to a generalization of max pooling in which the activation that stands out the most from the mean in each local neighborhood is selected, no matter whether it is a maximum or minimum.

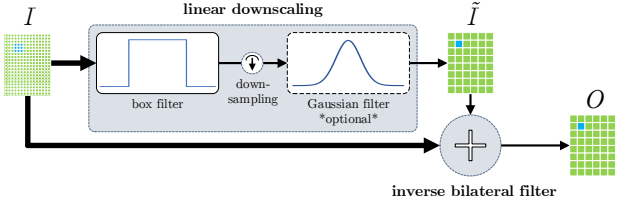

Figure 2. Diagram of detail-preserving downscaling (DPID) [31] and our detail-preserving pooling (DPP). DPP omits the Gaussian filter; Full-DPP replaces the box filter with a learned 2D filter.

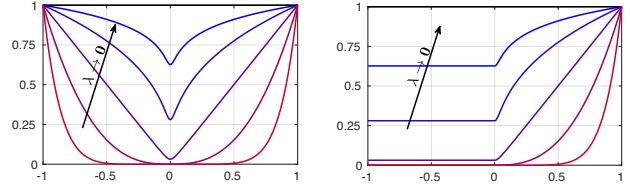

Figure 3. Inverse bilateral functions $\rho_\lambda(\cdot)$ used to calculate the weights of the input nodes in the neighborhood $\Omega_p$: The symmetric function (*left*) rewards arguments with bigger absolute values, the asymmetric function (*right*) rewards arguments with bigger values if they are positive. As smaller parameters $\lambda$ decrease the dynamic range of the reward function, it tends to being uniform for $\lambda \to 0$, leading to a simple averaging of all neighbors.

## 3. Detail-Preserving Image Downscaling

In contrast to traditional downscaling algorithms that aim for physically plausible results, more recent work has shown benefits of focusing on aspects of human perception. We specifically consider the approach of Weber *et al*. [31], which aims to preserve small details of the input image, which are often crucial for a faithful visual impression (Fig. 1). Their intuition is that small details transport more information than bigger areas with similar colors. To that end, an inverse bilateral filter is used to emphasize differences rather than punishing them. Given an input image $I[\cdot]$, detail-preserving image downscaling (DPID) calculates the downscaled output at pixel $p$ as

$$O[p] = \frac{1}{k_p} \sum_{q \in \Omega_p} I[q] \cdot \left\| I[q] - \tilde{I}[p] \right\|^\lambda, \qquad (1)$$

in which the linearly downscaled image $\tilde{I}$ is given by

$$\tilde{I} = I_D * \tfrac{1}{16} \begin{bmatrix} 1 & 2 & 1 \end{bmatrix}^T \begin{bmatrix} 1 & 2 & 1 \end{bmatrix}. \qquad (2)$$

$I_D$ is the result of a box filter applied to the input followed by downsampling, which is subsequently smoothed by an approximate 2D Gaussian filter. The weights are normalized with $k_p = \sum_{q \in \Omega_p} \| I[q] - \tilde{I}[p] \|^\lambda$. The neighborhood $\Omega_p$ is chosen according to the downscaling ratio. Downscaling by a factor of $k_H$ in one dimension and $k_W$ in the other results in applying Eq. (1) over a $k_W \times k_H$ neighborhood in the input image to form a single pixel in $O$. The overall block structure is illustrated in Fig. 2.

DPID calculates a weighted average of the input, but unlike ordinary bilateral filters [30], it *rewards* a difference in the input intensities such that pixel values with a bigger difference to $\tilde{I}$ contribute more significantly. This difference can be positive or negative, meaning that both darker and lighter pixels are rewarded based on their distance to $\tilde{I}$. This provides a customizable level of detail magnification, allowing to control the influence of regions with details, *e.g.* edges and corners. The parameter $\lambda$ adjusts the shape of the reward function, which can be tuned based on the image content and/or the downscaling ratio. A user study in [31] showed that on average people preferred DPID ($0.5 \leq \lambda \leq 1$) over all considered downscaling techniques.

## 4. Detail-Preserving Pooling

To apply the idea of detail-preserving image downscaling to CNNs, we here define a pooling layer called detail-preserving pooling (DPP). We propose two reward variants: a symmetric one that enhances all details and an asymmetric reward that only enhances details that stand out with a higher-than-average activation. To this end, we replace the $L_2$ norm in Eq. (1) with a generic scalar reward function, which will be learned, and apply all operations per channel. That is, we apply pooling to every feature map independently. Specifically, we define detail-preserving pooling of an input activation map $I$ at spatial output position $p$ as

$$\mathcal{D}_{\alpha,\lambda}(I)[p] = \frac{1}{\sum_{q' \in \Omega_p} w_{\alpha,\lambda}[p, q']} \sum_{q \in \Omega_p} w_{\alpha,\lambda}[p, q] I[q]. \qquad (3)$$

**Inverse bilateral weights.** Equation (3) computes a spatially weighted average of the input nodes in a neighborhood $I[q]_{q \in \Omega_p}$ for which we define weights $w_{\alpha,\lambda}[p, q]$ as

$$w_{\alpha,\lambda}[p, q] = \alpha + \rho_\lambda \left( I[q] - \tilde{I}[p] \right). \qquad (4)$$

The reward parameters $\alpha$ and $\lambda$ will be learned from data to enable the pooling to adapt to the requirements of each feature map. Note that we constrain the parameters to be non-negative by optimizing $\log \alpha$ and $\log \lambda$. For the symmetric variant of the reward function $\rho_\lambda(\cdot)$, we employ the differentiable (generalized) Charbonnier penalty [3, 28]

$$\rho_{\text{Sym}}(x) = \left( \sqrt{x^2 + \epsilon^2} \right)^\lambda \qquad (5)$$

with a small constant $\epsilon$. The asymmetric variant of $\rho_\lambda(\cdot)$ only rewards positive arguments and is formulated as

$$\rho_{\text{Asym}}(x) = \left( \sqrt{\max(0, x)^2 + \epsilon^2} \right)^\lambda, \qquad (6)$$

which prefers larger input activations by assigning bigger weights to nodes $q$ with $I[q] \geq \tilde{I}[p]$. In both cases, we occasionally omit the parameter $\lambda$ (referred to as the reward exponent) for the sake of notational simplicity. A bias term $\alpha$ is added to the weights to ensure that inputs from uniform regions are not entirely eliminated and influence the output.

For the sake of simplicity in notation, we reformulate this such that the weights are normalized as

$$\tilde{w}_{\alpha,\lambda}[p, q] = \frac{w_{\alpha,\lambda}[p, q]}{\sum_{q' \in \Omega_p} w_{\alpha,\lambda}[p, q']}, \quad (7)$$

which allows us to write DPP as

$$\mathcal{D}_{\alpha,\lambda}(I)[p] = \sum_{q \in \Omega_p} \tilde{w}_{\alpha,\lambda}[p, q] I[q]. \quad (8)$$

Depending on whether $\rho_{\text{Asym}}(x)$ or $\rho_{\text{Sym}}(x)$ is used for weight calculation, the final result is called *asymmetric* or *symmetric DPP*. Figure 3 visualizes the two families of weight functions. Note that we learn the shape of these nonlinear weight functions through their reward exponent $\lambda$. Figure 4 shows a pooling example. While from a pure visual impression, it is difficult to tell which type of pooling is better in the context of an entire deep network, our results below show that DPP clearly outperforms standard pooling layers even when placed deep in the network.

**Linear downscaling.** In Eq. (4), $\tilde{I}$ is the result of a linear downscaling. We achieve full flexibility with

$$\tilde{I}_F[p] = \sum_{q \in \tilde{\Omega}_p} F[q] I[q], \quad (9)$$

where $F$ is a learned, non-normalized 2D filter on regions $\tilde{\Omega}_p$. We typically use $3 \times 3$ regions $\tilde{\Omega}_p$, which is smaller than the linear filters used by [31]. This variant is referred to as *Full-DPP* in the following. Note that the filtering regions $\tilde{\Omega}_p$ can differ in size from neighborhoods in the inverse bilateral filter, *i.e.* $\Omega_p$; the pooling ratio of DPP is determined by the stride of downsampling following the linear filter $F$.

We also define a simplified variant termed *Lite-DPP* based on $\tilde{I}_{\text{Avg}}$, which is obtained with the special case of a non-learned box filter, *i.e.* $F[q] = 1/|\tilde{\Omega}_p|$, where $|\tilde{\Omega}_p|$ is the cardinality of $\tilde{\Omega}_p$. For Lite-DPP we use $2 \times 2$ filtering regions since we found no improvement from $3 \times 3$ regions.

**Learning and differentiability.** One drawback of max pooling is that it is not differentiable, hence one has to resort to a sub-differential during learning. A look-up table is used for backpropagation, which is computed and stored during the forward pass. The consequence is that the gradient is only flowing to the maximizer during the backward pass, which can be problematic when the network is deeper. Note that pooling layers that include max pooling as a component, *e.g.* by learning a linear combination of max and average pooling [19], typically inherit this non-differentiability. DPP, on the other hand, is fully differentiable:

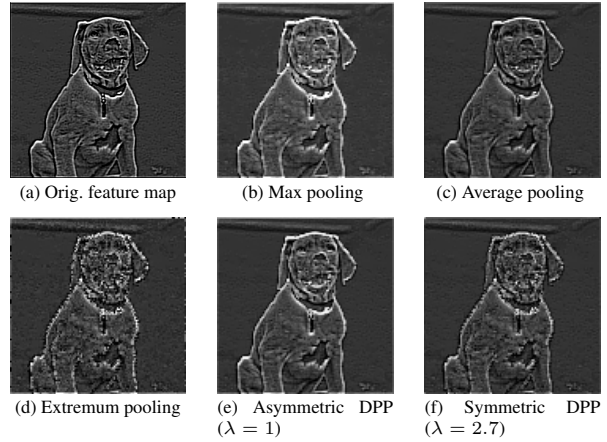

(a) Orig. feature map  (b) Max pooling  (c) Average pooling

(d) Extremum pooling  (e) Asymmetric DPP ($\lambda = 1$)  (f) Symmetric DPP ($\lambda = 2.7$)

Figure 4. Visualization of different pooling methods on an example feature map taken from the second layer of VGG-16. For both reward variants, the bias $\alpha$ is set to 0 to visually magnify the effect of the inverse bilateral weights. *Best viewed on screen.*

**Proposition 1.** $\mathcal{D}_{\alpha,\lambda}$ *is differentiable w.r.t. $I$, $\alpha$, and $\lambda$.*

*Proof.* For $\alpha$ this is obvious. For $\lambda$ and $I$ this follows directly from $\rho_\lambda$ being differentiable. In the asymmetric case, the derivative of $\rho_{\text{Asym}}$ w.r.t. $x$ is

$$\frac{d\rho_{\text{Asym}}(x)}{dx} = \begin{cases} 0 & x \leq 0 \\ \lambda x (x^2 + \epsilon^2)^{\frac{\lambda}{2}-1} & 0 < x. \end{cases} \quad \square \quad (10)$$

Leveraging this differentiability characteristic, we let the network learn the shape of the weight function by learning both $\alpha$ and $\lambda$ and, optionally, the filter parameters $F$ in linear downscaling. We initialize the parameters of the inverse bilateral function to 1 and the filter coefficients of the linear downscaling to 0. A small zero-mean Gaussian perturbation is added to all parameters to break any undesired symmetries. The standard deviation is set as suggested by [6]; $\epsilon^2$ is set to 0.001. We use stochastic gradient descent (SGD) for training with weight decay disabled for $\alpha$ and $\lambda$.

**Stochastic variants.** Stochastic pooling techniques [33, 34] fuse downsampling and regularization through a random node selection procedure to improve performance. Stochastic pooling [33] randomly selects one node in every $s \times s$ neighborhood, where $s$ is the pooling factor, according to a multinomial distribution that favors nodes with stronger activations. S3pool [34] applies $s \times s$ max pooling with a stride of 1 and then applies uniform sampling to rows and columns of nodes. Inspired by this, we can also define stochastic variants of DPP. Specifically, we consider *stochastic spatial sampling DPP* (S3DPP), which extends S3Pool by first applying regular DPP with a stride of 1 to the input and then sampling rows and columns uniformly. Since DPP magnifies details in a feature map, adding a uniform random selection of nodes on top results in a controlled importance sampling where the importance of every node is controlled by the DPP parameters $\lambda$ and $\alpha$.

## 5. Analysis and Discussion

We turn to a discussion and brief mathematical analysis of certain key properties of the proposed DPP layer. Particularly, we show how DPP can learn to behave similar to other pooling layers. We also discuss its computational and parameter overhead, as well as some motivations.

**Relation to other pooling layers.** First, we mathematically show that symmetric DPP can be equivalent to average or extremum (max in the case of asymmetric DPP) pooling for certain choices of parameters.

**Proposition 2.** $\mathcal{D}_{\alpha,0}$ *is equivalent to average-pooling for any finite* $\alpha \in \mathbb{R}^+$.

*Proof.* $\lambda = 0$ (and finite, positive $\alpha$) implies that the weights of Eq. (4) are equal. Hence, the normalized weights (Eq. 7) equal $1/|\Omega|_p$. Consequently, Eq. (8) performs averaging of all activations in the neighborhood $\Omega_p$. □

**Proposition 3.** *As* $\lambda \to \infty$*, symmetric* $\mathcal{D}_{0,\lambda}$ *performs extremum pooling while asymmetric* $\mathcal{D}_{0,\lambda}$ *yields max pooling.*

*Proof.* We begin with the symmetric case. Let $q_e$ be the location of the extremum of the neighborhood, *i.e.*

$$q_e = \arg\max_{q \in \Omega_p} |I(q) - \tilde{I}(p)|. \tag{11}$$

We first consider $\tilde{w}_{0,\lambda}[q_e]$, omitting the argument $p$ for brevity. After rewriting Eq. (7) for the extremum pixel and dividing the numerator and denominator by $w_{0,\lambda}[q_e]$, we have

$$\frac{1}{1 + \sum_{q' \neq q_e} w_{0,\lambda}[q']/w_{0,\lambda}[q_e]}. \tag{12}$$

Since $q_e$ is the location of the extremum and assuming no ties, $w_{0,\lambda}[q']/w_{0,\lambda}[q_e] \to 0$ for $\lambda \to \infty$. It follows that $\lim_{\lambda \to \infty} \tilde{w}_{0,\lambda}[q_e] = 1$.

Next, we consider all non-extremum pixels, *i.e.* $\tilde{w}_{0,\lambda}[q], q \neq q_e$, which we can rewrite as

$$\frac{1}{w_{0,\lambda}[q_e]/w_{0,\lambda}[q] + \sum_{q' \neq q_e} w_{0,\lambda}[q']/w_{0,\lambda}[q]}. \tag{13}$$

Reasoning as above leads to $w_{0,\lambda}[q_e]/w_{0,\lambda}[q] \to \infty$ for $\lambda \to \infty$. Given that all $w_{0,\lambda}[q']/w_{0,\lambda}[q] \geq 0$, it follows that $\lim_{\lambda \to \infty} \tilde{w}_{0,\lambda}[q] = 0$. In the case of ties, the extrema all receive equal weight, leading to their averaging as $\lambda \to \infty$.

For the asymmetric case, the reasoning is analogous, except that we omit the absolute value in Eq. (11). □

**Parameterization.** DPP is applied to all channels independently, hence results in adding one $\lambda$ and one $\alpha$ parameter per feature map. If block averaging is used for linear downscaling, no further parameters are added. The full $\tilde{I}_F$ from Eq. (9), using $3 \times 3$ filters, adds 10 parameters (including the bias in convolutions) per feature map. Note that this is by far less (2 or 3 orders of magnitude, depending on the number of channels) than a single 3D convolutional layer, which would add $(k_W \times k_H \times nP \times nP) + nP$ parameters to the network, where $k_W$ and $k_H$ are the width and height of the filter (here, 3) and $nP$ is the number of feature maps.

ResNet-50 and 101 have 25M and 44M parameters, VGG-16 has 139M. Full-DPP with learnable linear filters adds 43k parameters (0.172% and 0.098%) to ResNet and 17.7k parameters to VGG-16 (0.013%). For Lite-DPP with fixed linear downsampling, as used for most experiments, the increase in parameter count is even more negligible.

**Computational expense.** The proposed detail-preserving pooling is conveniently scalable since every pixel in the output is independent of the others. Hence a parallel implementation achieves a considerable speed-up. Since the linear downscaling part of DPP is using standard layers, we only require a CUDA implementation of the nonlinear component. Code is available on Github [35].

The inverse bilateral weight calculation from Eq. (3), on average, takes 45ms on a single Pascal Titan X GPU to perform a forward and backward pass for a mini batch of 128 images, 64 feature maps, $224 \times 224$ spatial resolution, and a pooling ratio of 2. This is the worst case overhead for the majority of existing CNN architectures. The overhead depends on the number of pooling layers, but is independent of the number of convolutional layers. The slow-down of DPP for VGG-16 on ImageNet is $\sim 20\%$, while for very deep networks it is quite minor, *e.g.* $\sim 5\%$ for ResNet-101.

**Detail preservation in deeper layers.** While a perceptual motivation of DPP is immediate in earlier network layers, the activations of deeper layers are visually rather distinct from natural images (Fig. 4). To study the benefit of DPP in deeper layers, we trained a ResNet-110 [11] on the CIFAR10 dataset [16] with data augmentation (see Sec. 6 for details). We compare three variants: *(1)* a standard ResNet-

| Network | Error [%] |
|---|---|
| ResNet-110 (standard, $2\times$ strided convolution) | 6.89 |
| ResNet-110 ($1\times$ strided convolution + $1\times$ DPP) | 6.68 |
| ResNet-110 ($2\times$ DPP) | 6.59 |

Table 1. Impact of replacing the downsampling steps of ResNet [11] with DPP, evaluated on CIFAR10. Replacing only the second downsampling with DPP (2$^{nd}$ row) already yields a clear improvement, showing the benefit of DPP even in deeper layers. Replacing both downsampling steps with DPP further improves the results.

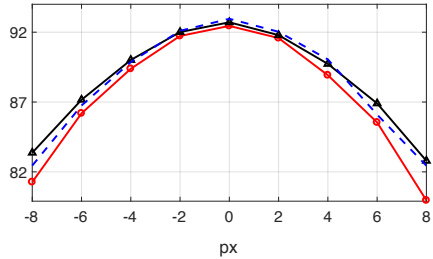

Figure 5. Classification accuracy (in % using VGG on CIFAR10) as a function of horizontal input translations (in pixels). DPP is more translation-invariant than max pooling, and performs best for moderate translations of up to 4 pixels (12.5% of the image size).

110, *(2)* ResNet where only the second (deeper) downsampling has been replaced by DPP, and *(3)* ResNet in which both strided convolutions have been replaced with DPP. Table 1 shows the results. Having just one DPP located in a very deep layer clearly improves the accuracy, illustrating the benefit of DPP even for deeper network layers.

**Invariance to small perturbations.** One of the benefits of pooling is gaining invariance to small perturbations in the input image, such as translations or rotations. As an example analysis, Fig. 5 compares the robustness of DPP to that of max and average pooling when subjected to a horizontal shift of the input. The experiment is performed using a VGG [27] network on CIFAR10. DPP shows better translation invariance compared to max pooling, and outperforms average pooling for moderate amounts of translation (up to 4 pixels). DPP thus shows a favorable trade-off between invariance and classification accuracy (*c.f.* Sec. 6).

## 6. Experiments

We verify the effectiveness of our approach with a series of experiments on different datasets and CNN architectures. First, we perform image classification on the CIFAR10 dataset [16] as this allows for exhaustive comparisons. The trained networks include ResNet-110 [11], VGG [27], a NIN-like network [20], and DenseNet [12]. The original VGG uses max pooling, while NIN uses average and max pooling both in the same network. ResNet-110 employs strided convolutions for downscaling, while DenseNet relies on average pooling. We will show that in some cases, *e.g.* VGG and NIN on CIFAR10, the original pooling layer is suboptimal and can be improved upon with other standard pooling methods. Such a procedure is of course tedious, possibly even impractical depending on the application dataset and network. Moreover, for all four architectures, our pooling layer clearly learns to perform better than the original choice by a considerable margin. Compared to other, more advanced pooling methods, DPP also shows favorable results. The second set of experiments focuses on using bigger networks on a more realistic dataset. We train complete VGG-16 and ResNet-50 and 101 networks on the classification task of the ImageNet dataset [25] and observe clear improvements over the original networks.

We test two linear downscaling variants inside DPP, *Lite-DPP* and *Full-DPP*. Moreover, we compare asymmetric (*Asym*) and symmetric (*Sym*) reward functions. By default we use a pooling ratio of 2, *i.e.* set the downsampling stride after the linear filter to 2, and apply the inverse bilateral filter on non-overlapping $2 \times 2$ neighborhoods $\Omega_p$. Finally, we consider the stochastic *S3DPP* in some of the settings. Note that for all networks and experiments, the respective parameters of DPP are learned from data; every feature channel has unique parameters, which are not fixed.

### 6.1. Detailed analysis on CIFAR10

The CIFAR10 dataset [16] consists of 60000 images ($32 \times 32$ pixels, RGB) evenly distributed across 10 classes. 50000 images are used for training, 10000 for testing. Stochastic gradient descent is used for optimization, with momentum set to 0.9 and a mini-batch size of 128. The initial learning rate is set to 1 and cut in half every 25 epochs. The whole training procedure lasts 300 epochs. Apart from commonly used pooling layers, we also compare against mixed (50/50) and gated pooling [19], as well as $L_p$ pooling [2]. We did not include tree pooling [19] since it was outperformed by mixed and gated pooling on CIFAR10 in the original paper. We additionally consider stochastic methods, particularly stochastic pooling [33] and S3pool [34].

**VGG.** We first use a network with an architecture similar to VGG-16 [27], but with smaller fully connected layers to reduce the parameter count (two fully connected layers of 512 and 10 nodes). Each convolution layer is followed by a batch normalization and a ReLU nonlinearity. We augment the dataset during training by flipping the input images with a probability of 0.5. Our results are the average of 10 runs.

**NIN.** This network is based on the idea of using $1 \times 1$ convolutional layers, follows a very different structure compared to VGG, and has considerably fewer parameters. We employ a configuration very similar to that of [20]. A fully connected layer is added to the end of the network to aid convergence (see supplemental). The data is preprocessed as in [20]; we perform 3 trials without data augmentation.

**ResNet.** Next, we use ResNet-110 [11], a very deep network, which has been shown to have the best accuracy among all ResNets on CIFAR10. ResNet consists of numerous concatenated residual building blocks. Unlike VGG and NIN, it does not have layers explicitly dedicated to pooling, but instead sets the stride of a convolution inside some of the building blocks to 2 to achieve downscaling. ResNet-110 for CIFAR10 downscales the input in two of its building blocks. For our experiments we set the stride of these two blocks back to 1 and place different pooling layers (max, average, DPP) immediately after each of them, while leaving the rest of the network fully intact. We also leave the global average pooling unchanged. Each experiment (for all pooling types) is repeated 5 times and the best result is reported, again following the standard protocol [11].

| | Method | VGG | NIN | ResNet |
|---|---|---|---|---|
| Deterministic methods | Strided conv. | 8.43±0.20 | 10.97±0.10 | 6.23$^{(*)}$ |
| | Max | 7.43±0.20$^{(*)}$ | 9.42±0.07 | 6.52 |
| | Average | 7.12±0.18 | 8.75±0.15 | 6.33 |
| | NIN | – | 9.01±0.11$^{(*)}$ | – |
| | Mixed (50/50) | 7.27±0.20 | 8.68±0.23 | 6.05 |
| | Gated | 7.25±0.14 | 8.67±0.22 | 7.12 |
| | $L_2$ | 7.15±0.18 | 8.65±0.12 | 7.29 |
| | Lite-DPP$_{Asym}$ | 7.10±0.15 | 8.62±0.10 | 6.17 |
| | Full-DPP$_{Asym}$ | 7.17±0.18 | 8.73±0.05 | 6.23 |
| | Lite-DPP$_{Sym}$ | 7.19±0.10 | 8.58±0.11 | 6.05 |
| | Full-DPP$_{Sym}$ | **7.02**±0.18 | 8.70±0.14 | 5.97 |
| Stoch. | Stochastic | 7.67±0.10 | 8.92±0.09 | 5.83 |
| | S3pool | 7.21±0.14 | 7.23±0.08 | 5.55 |
| | Lite-S3DPP$_{Sym}$ | – | **7.13**±0.09 | **5.42** |

Table 2. Comparison of different architectures and pooling layers on the CIFAR10 dataset (classification error in %, best result bold, 2$^{nd}$ best underlined). For VGG and NIN we report the mean and standard deviation across trials, for ResNet the error for the best model following [11]. The original choice of pooling layers (marked as ∗) is not necessarily optimal for CIFAR10. DPP outperforms the best standard pooling (top part) in all cases, and also exceeds more advanced pooling layers (2$^{nd}$ part). Moreover, DPP yields the best results among the stochastic methods (bottom part).

**DenseNet.** Finally, we apply S3DPP to the more recent DenseNet [12] architecture. For this experiment we take DenseNet-BC ($L = 100$, $k = 24$) and replace the average poolings in the transition layers with Lite-S3DPP$_{Sym}$ and train (mini-batch size 64) with data augmentation. We report results from a single run, following [12].

**Discussion.** The comparison to standard pooling methods in Table 2 (top part) for all experiments reveals that none of strided convolution, max, and average pooling, or their combination are consistently superior. In fact, which is best can be different from the original choice of the respective network, which was typically optimized for a specific dataset. This clearly demonstrates the necessity of a generic pooling layer that can perform better than the best baseline.

Linear combinations of max and average pooling, e.g. the 50/50 mixed or gated pooling [19], are observed to perform somewhere between max and average for VGG, yield some improvements for NIN, and show inconsistent results on ResNet. Unlike the settings in [19], mixed pooling does not outperform the baselines in all cases.

Spatial $L_p$ pooling [9] calculates the $p$-norm over input neighborhoods. Finding the best $p$ following the approach of [26] requires brute force evaluation across a wide range of $p$ values on a validation set. This is computationally expensive and limits practicality.[2] We choose $L_2$ pooling for comparison, as it is the most commonly used variant. It outperforms the baseline methods for NIN, but significantly deteriorates accuracy on ResNet.

| Network | Params [M] | Error [%] |
|---|---|---|
| DenseNet-BC | 3.020 | 3.90 |
| DenseNet | 27.2 | 3.74 |
| DenseNet-BC + Lite-S3DPP$_{Sym}$ | 3.021 | 3.75 |

Table 3. Comparison of the effect of adding Lite-S3DPP$_{Sym}$ to DenseNet ($k = 24$) vs. increasing the number of feature maps in the transition layers. Replacing average pooling with S3DPP adds 1032 parameters. The same error decrease as from S3DPP can be obtained had we used regular DenseNet (not the BC version), at the expense of a 9-fold increase in parameter count.

Comparing to all deterministic pooling methods, DPP shows strong results, yielding the *lowest error for all three networks*. For VGG, the number of trials enables establishing statistical significance. For example, we can confidently reject the hypothesis (with 99% confidence) that the difference of DPP to max pooling and 50/50 mixed pooling is due to chance. While all DPP variants perform well, the symmetric ones tend to perform best and reach 5.97% error using ResNet-110 on CIFAR10. This is perhaps unexpected, since the limit case of symmetric rewards for large values of $\lambda$ is extremum pooling, a concept that had not been considered so far. One important property of DPP is that, unlike all other pooling methods tested, it achieves *consistently* good results across three different networks (and also other datasets, see below). Hence, we suggest that DPP can be used as a generic pooling layer, which can avoid finding the best pooling layer using brute force enumeration. Following our analysis, we recommend Lite-DPP$_{Sym}$ as default choice and Full-DPP$_{Sym}$ if computation is not critical.

Among the stochastic pooling layers (Table 2, bottom), the variant of DPP consistently performs the *best* as well, yielding very competitive results for ResNet (5.42% error). Favorable results can also be observed with DenseNet in Table 3, showing that the lean DenseNet-BC with S3DPP becomes competitive with the much larger standard DenseNet (no -BC) with many times more parameters. Note, however, that too much stochasticity can limit the accuracy [34].

Figure 6 shows the distribution of the learned pooling parameters in the different pooling layers of a VGG network with Lite-DPP$_{Asym}$ for CIFAR10. For higher values of $\alpha$, the reward bias becomes the dominant term and the pooling layer performs similar to average pooling. With low $\alpha$ val-ues, as in the early layers, the inverse bilateral behavior becomes the dominant characteristic of DPP. Thereby, higher values of the reward exponent $\lambda$ correspond to a behavior more similar to max pooling. We observe that lower layers tend to behave more like max pooling, while later layers average more. Other networks exhibit similar trends.

---
[2]Learning the $p$ value would require a modified formulation due to possible exploding gradients.

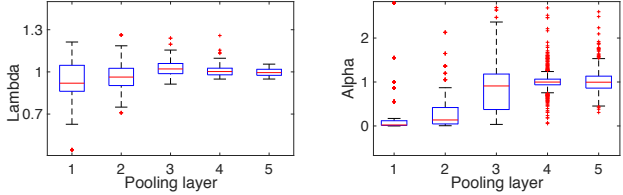

Figure 6. Empirical distribution of learned $\lambda$ (*left*) and $\alpha$ (*right*) for different pooling layers in a VGG network with Lite-DPP$_{\text{Asym}}$ on CIFAR10. Higher values of $\lambda$ indicate a more max-like behavior, while higher values of $\alpha$ indicate an averaging tendency.

### 6.2. Image classification on ImageNet

The ILSVRC dataset [25] has 1.2M images for training and 50k for validation, distributed evenly across 1000 classes. We use $224 \times 224$ random crops (after rescaling) for training and test on the validation set.

**VGG-16.** We train a standard VGG-16 network [27] to yield a baseline, and compare this to the case where the original max pooling is swapped for the proposed DPP. For completeness, we also report results using average pooling. All networks are trained from scratch and as similar to the original training conditions of [27] as possible. We add 2 batch normalization units after the 1$^{\text{st}}$ and the 2$^{\text{nd}}$ fully connected layers to help reduce the sensitivity to initialization. Both networks are initialized as suggested by [6] and trained for 74 epochs, using mini batches of 128 images. Baseline training took 1 hour per epoch on 4 Pascal Titan X GPUs and the added training time caused by our pooling layer was no more than 12 minutes per epoch. The relative computation overhead for VGG was the worst case scenario among all our ImageNet experiments.

**ResNet.** We also evaluate on ResNet-50 and 101. For ImageNet, there are three residual blocks that perform downscaling and, as before, we set the stride in these blocks back to 1 and place DPP immediately after each block. The preprocessing of the data and training are done as in [11] for 90 epochs. We use a mini-batch of size 85 for training and we also train the baseline networks with the same batch size to maintain comparability. The initial learning rate is set to 0.033 and cut by a factor of 10 every 30 epochs.

**Results.** For both networks we observe that using DPP as pooling layer accelerates convergence compared to the standard architecture especially at the early training stages. More importantly, our VGG-16 converges to 30.05% error for the one crop evaluation criterion, thus 0.3% points lower than the baseline with max pooling, which converges to 30.4% error. For ResNet the gap is wider. Table 4 shows a detailed comparison for both networks. It should be noted that our ResNet baselines are already better than the numbers reported in [11], which were 22.85% and 21.75% for the ten-crop evaluation of ResNet-50 and 101. Adding DPP lowers the error on the validation set by 0.8% for ResNet-50 and by 0.6% points for ResNet-101. In fact, ResNet-101 with DPP surpasses a standard ResNet-152 with quite a bit of margin, despite being faster (5 *vs.* 50% slow-down over ResNet-101) and having many fewer layers and parameters.

| Network | Single crop | | Ten crop | |
|---|---|---|---|---|
| | Top-1 | Top-5 | Top-1 | Top-5 |
| VGG-16 (original) | 30.40 | 10.20 | – | – |
| VGG-16 + Average | 30.36 | 10.19 | – | – |
| VGG-16 + Lite-DPP$_{\text{Sym}}$ | **30.05** | **10.08** | – | – |
| ResNet-50 | 24.23 | 7.26 | 22.53 | 6.26 |
| ResNet-50 + Lite-DPP$_{\text{Sym}}$ | **23.42** | **6.83** | **21.82** | **5.92** |
| ResNet-101 | 22.31 | 6.23 | 20.93 | 5.38 |
| ResNet-152 | 22.16 | 6.16 | 20.69 | 5.21 |
| ResNet-101 + Lite-DPP$_{\text{Sym}}$ | **21.70** | **5.91** | **20.52** | **5.20** |

Table 4. Comparison of VGG-16, ResNet-50 and 101 (error in %) on ImageNet, with and without DPP for different evaluation methods. DPP yields consistent benefits. Our ResNet-101 with DPP even outperforms the much larger and deeper ResNet-152.

## 7. Conclusion

We presented a novel pooling layer for convolutional neural networks termed detail-preserving pooling (DPP), based on the idea of inverse bilateral filters. DPP allows downscaling to focus on important structural detail; learnable parameters control the amount of detail preservation. We showed theoretically that DPP can adapt to perform similar to max/extremum or average pooling, or on a nonlinear continuum of intermediate functions while incurring only a minor computational overhead. Our quantitative experiments showed that for a wide range of network architectures and datasets, DPP performs consistently better than the best standard pooling layer and a selection of advanced pooling methods, making DPP broadly applicable. Furthermore, DPP can be combined with stochastic pooling methods with further accuracy gains as detail preservation and regularization complement each other. For ResNet on the challenging ImageNet classification task, DPP substantially lowers the error, even below that of much deeper models.

**Acknowledgements.** FS and SR gratefully acknowledge support by Smiths Heimann GmbH. The work of NW is supported by the 'Excellence Initiative' of the German Federal and State Governments and the Graduate School of Computational Engineering at TU Darmstadt.

# Detail-Preserving Pooling in Deep Networks
## – Supplemental Material –

Faraz Saeedan[1]    Nicolas Weber[1,2][*]    Michael Goesele[1,3][*]    Stefan Roth[1]

[1]TU Darmstadt    [2]NEC Laboratories Europe    [3]Oculus Research

## A. Learned Pooling Parameters

The learned pooling parameters for a VGG network with Lite-DPP$_{\text{Asym}}$ after training on CIFAR10 have been shown in Fig. 6 in the main manuscript. Here, Fig. 7 additionally shows the learned pooling parameters for a VGG network using the *symmetric* Lite-DPP$_{\text{Sym}}$, trained on the same dataset. It can be seen that the general trend of the pooling parameters resembles that of the asymmetric case. The earlier pooling layers have a smaller reward bias, which indicates an extremum-like tendency, while the deeper layers learn larger bias values and hence behave more similar to average pooling. Compared to the asymmetric case, here the reward exponent learned for the early layers is smaller, which indicates that the network seems to prefer a somewhat smaller level of detail preservation with extremum pooling compared to max pooling in the asymmetric case.

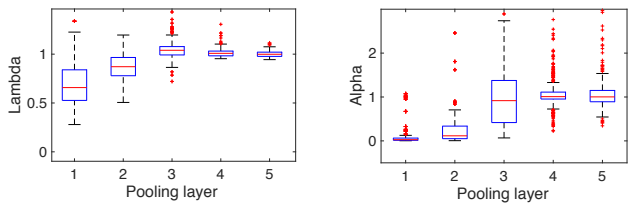

Figure 7. Illustration of the learned $\lambda$ (*left*) and $\alpha$ (*right*) for different pooling layers in a VGG network with Lite-DPP$_{\text{Sym}}$ on CIFAR10. Higher values of $\lambda$ indicate a more extremum-like behavior, while higher values of $\alpha$ indicate an averaging tendency.

## B. Network Details

For our experiments on CIFAR10, we have used a VGG-like and a NIN-like network (aside from RestNet-110 and DenseNet-BC). The exact architecture of the former two networks is given in Table 5. Every convolution layer is followed by a batch normalization and a ReLU nonlinearity. The NIN-like network has a fully connected layer after the global average pooling, as we found that adding this layer helps avoiding the otherwise fairly common problem of local minima.

To apply different pooling methods we have simply swapped the pooling layers in the original networks of Table 5 with the desired pooling layers. To use strided convolutions instead, we have removed the pooling layers and instead set the stride in the convolution layers before the original pooling locations to two.

| VGG configuration | NIN configuration |
|---|---|
| conv3-64 $\times 2$ | conv5-192 $\times 1$ |
| **pooling/2** | conv1-160 $\times 1$ |
| conv3-128 $\times 2$ | conv1-96 $\times 1$ |
| **pooling/2** | **pooling/2** |
| conv3-256 $\times 3$ | conv5-192 $\times 1$ |
| **pooling/2** | conv1-192 $\times 2$ |
| conv3-512 $\times 3$ | |
| **pooling/2** | **pooling/2** |
| conv3-512 $\times 3$ | conv3-192 $\times 1$ |
| **pooling/2** | conv1-192 $\times 2$ |
| FC-512 | avg pooling/8 |
| FC-10 | FC-10 |

Table 5. The configurations of two of the networks used for the CIFAR10 experiments. Each convolutional layer is followed by batch normalization and a ReLU unit (not shown). The layers in bold have been replaced with DPP.

---
[*]This work was carried out while at TU Darmstadt.